\documentclass[submission,copyright,creativecommons]{eptcs}
\providecommand{\event}{ICLP DC 2020} 
\usepackage{breakurl}             
\usepackage{underscore}           

\usepackage[english]{babel}
\usepackage{times}
\usepackage{helvet}
\usepackage{courier}
\usepackage{amsmath}
\usepackage{amssymb}
\usepackage{enumerate}
\usepackage{graphicx}
\usepackage{url}
\usepackage{listings}
\usepackage{algorithm}

\def\mvis{\!=\!}

\def\bi{\begin{itemize}}

\def\ei{\end{itemize}}
\def\beq{\begin{equation}}
\def\eeq#1{\label{#1}\end{equation}}
\def\ba{\begin{array}}
\def\ea{\end{array}}
\def\i#1{\hbox{\it #1\/}}

\def\lpmln{\hbox{\rm LP}^{\rm{MLN}}}
\def\lpmln{{\rm LP}^{\rm{MLN}}}

\def\sneg{\sim\!\!}
\def\ar{\leftarrow}

\def\false{\hbox{\sc false}}
\def\true{\hbox{\sc true}}

\def\i#1{\hbox{\itshape #1\/}}

\def\nasp{{\rm NeurASP}}

\newtheorem{example}{Example}

\long\def\BOC#1\EOC{\message{(Commented text )}}
\long\def\BOCC#1\EOCC{\message{(Commented text )}}
\long\def\BOCCC#1\EOCCC{\message{(Commented text )}}
\long\def\optional#1{\empty}
\long\def\NB#1{}
\long\def\NBB#1{}

\title{Extending Answer Set Programs with Neural Networks} 
\author{
Zhun Yang
\institute{ 
Arizona State University, Tempe, AZ, USA}
\email
{zyang90@asu.edu}
}

\def\titlerunning{$\nasp$}
\def\authorrunning{Yang}

\begin{document}
\maketitle

\begin{abstract}
The integration of low-level perception with high-level reasoning is one of the oldest problems in Artificial Intelligence. Recently, several proposals were made to implement the reasoning process in complex neural network architectures. While these works aim at extending neural networks with the capability of reasoning, a natural question that we consider is: can we extend answer set programs with neural networks to allow complex and high-level reasoning on neural network outputs?
As a preliminary result, we propose 
$\nasp$ --- a simple extension of answer set programs by embracing neural networks where neural network outputs are treated as probability distributions over atomic facts in answer set programs.
We show that $\nasp$ can not only improve the perception accuracy of a pre-trained neural network, but also help to train a neural network better by giving regularization through logic rules. However, training with $\nasp$ implementation takes much more time than pure neural network training due to the internal use of a symbolic reasoning engine.
For future work, we plan to investigate the potential ways to solve the scalability issue of $\nasp$ implementation. One way is to embed logic programs directly in neural networks. On this route, we plan to design a SAT solver using neural networks and extend such a solver to allow logic programs.


\end{abstract}

\section{Introduction and Problem Description}
The integration of low-level perception with high-level reasoning is one of the oldest problems in Artificial Intelligence. This topic is revisited with the recent rise of deep neural networks. Several proposals were made to implement the  reasoning process in complex neural network architectures, e.g., \cite{kathryn18tensorlog,rocktaschel17end,donadello17logic,kazemi18relnn,sourek15lifted,palm18recurrent,lin19kagnet}. 
However, it is still not clear how complex and high-level reasoning, such as default reasoning \cite{rei80}, ontology reasoning \cite{baader03handbook}, and causal reasoning \cite{pearl00causality}, can be successfully computed by these approaches. 
The latter subject has been well-studied in the area of knowledge representation (KR), but many KR formalisms, including answer set programming (ASP) \cite{lif08,bre11}, are logic-oriented and do not incorporate high-dimensional vector space and pre-trained models for perception tasks as handled in deep learning, which limits the applicability of KR in many practical applications involving data and uncertainty.
A natural research question we consider is: can we extend answer set programs with neural networks to allow complex and high-level reasoning on information provided in vector space?

Our research tries to answer this question. 
To start with, we extended $\lpmln$, a probabilistic extension of ASP, with neural networks by turning neural network outputs into weighted rules in $\lpmln$. However, there is a technical challenge: the existing parameter learning method of $\lpmln$ is too slow to be coupled with typical neural network training. This motivates us to consider a simpler probabilistic extension of ASP, for which we design and implement an efficient parameter learning method.

In this research summary, we present our preliminary work --- $\nasp$, which is a simple and effective way to integrate sub-symbolic and symbolic computation under stable model semantics while the neural network outputs are treated as the probability distribution over atomic facts in answer set programs. 
We demonstrate how $\nasp$ can be useful for some tasks where both perception and reasoning are required, and
show that $\nasp$ can not only improve the perception accuracy of a pre-trained neural network, but also help to train a neural network better by giving restrictions through ASP rules.

The biggest issue we are encountering now is that training with $\nasp$ implementation still takes much more time than pure neural network training due to the internal use of a symbolic reasoning engine (i.e., {\sc clingo} in our case). We plan to investigate two directions to resolve this issue. One is to compute $\nasp$ in a circuit. 
For example, we can turn an $\nasp$ program into a Probabilistic Sentential Decision Diagram (PSDD) \cite{kisa14probabilistic} so that the probability and gradient computation would take linear time in every iteration. The challenge for this route is how to construct a circuit efficiently. The other direction is to embed the logic reasoning part completely in neural networks without referring to any symbolic reasoning engines. 
The challenge is how to embed logic rules in neural networks while maintaining the expressivity.

Besides, we plan to apply $\nasp$ to domains that require both perception and reasoning. The first domain we are going to investigate is visual question-answering and we limit our attention to the problems whose reasoning can be potentially represented in ASP. Some well-known datasets in this domain include NLVR2 \cite{suhr18corpus}, CLEVR \cite{johnson17clevr}, and CLEVRER \cite{yi19clevrer}. We plan to start with replacing the reasoning part in existing works with $\nasp$ and analyze the pros and cons of applying $\nasp$ on those visual reasoning domains. We also plan to apply $\nasp$ to predict whether a vulnerability will be exploited so that the knowledge from domain experts can help neural network training.

The paper will give a summary of my research, including some background knowledge and reviews of existing literature (Section 2), goal of my research (Section 3), the current status of my research (Section 4), the preliminary results we accomplished (Section 5), and some open issues and expected achievements (Section 6).
The implementation of our prelimilary work --- $\nasp$, as well as codes used for the experiments, is publicly available online at 
\begin{center}
\url{https://github.com/azreasoners/NeurASP}.\par
\end{center}

\section{Background and Overview of the Existing Literature}
Recent years have observed the rising interests of combining perception and reasoning.

DeepProbLog \cite{manhaeve18deepproblog} extends ProbLog with neural networks by means of neural predicates. We follow similar idea to design $\nasp$ to extend answer set programs with neural networks. Some differences are: (i) The computation of DeepProbLog relies on constructing an SDD whereas we use an ASP solver internally. (ii) $\nasp$ employs expressive reasoning originating from answer set programming, such as defaults, aggregates, and optimization rules. This not only gives more expressive reasoning but also allows the more semantic-rich constructs as guide to learning. (iii) DeepProbLog requires each training data to be a single atom, while $\nasp$ allows each training data to be arbitrary propositional formulas.

Xu et al. \cite{xu18asemantic} used the semantic constraints to train neural networks better, but the constraints used in that work are simple propositional formulas whereas we are interested in answer set programming language, in which it is more convenient to encode complex KR constraints.
Logic Tensor Network \cite{donadello17logic} is also related in that it uses neural networks to provide fuzzy values to atoms.

\BOC
Several other proposals were made to integrate statistical models and symbolic knowledge through loss functions, which determine the directions towards which the parameters of the statistical model are updated. \cite{hu16harnessing} proposed a general framework of using logic rules to enhance neural networks, where the neural network is seen as a ``student,'' while a ``teacher network'' uses a set of weighted first-order rules over the input-target space to regulate the prediction from the student network. The loss function does not only consider the distance between the prediction from the student network and the ground truth, but also the distance between predictions from the student network and from the teacher network. In this way, the student network is trained not only to fit the training data but also simulate the behavior of the teacher network, which is operated by a set of logic rules. Similarly, \cite{xu18asemantic} has investigated a semantic loss function that bridges between the vector output from neural networks and how much a given set of logical constraints are satisfied by the vector output. 
In such a loose integration of logic rules and neural network through loss functions, the logical component is often seen as a blackbox --- the overall training process of the neural network is agnostic to the exact syntax and semantics of the logical component. 
\EOC 

Another approach 
is to embed logic rules in neural networks by representing logical connectives by mathematical operations and allowing the value of an atom to be a real number. For example, 
Neural Theorem Prover (NTP) \cite{rocktaschel17end} adopts the idea of dynamic neural module networks \cite{andreas16learning} to embed logic conjunction and disjunction in and/or-module networks. A proof-tree like end-to-end differentiable neural network is then constructed using Prolog's backward chaining algorithm with these modules. 
Another method that also constructs a proof-tree like neural network is TensorLog \cite{kathryn18tensorlog}, which uses matrix multiplication to simulate belief propagation that is tractable under the restriction that each rule is negation-free and can be transformed into a polytree. 

Graph neural network (GNN) \cite{kipf17semi} is a neural network model that is gaining more attention recently. Since a graph can encode objects and relations between objects, by learning message functions between the nodes, one can perform certain relational reasoning over the objects. For example, in~\cite{palm18recurrent}, it is shown that GNN can do well on Sudoku, but the input there is not an image but a textual representation. 
Another work NeuroSAT \cite{selsam18learning} shows that GNN can solve general small SAT problems after only being trained as a classifier to predict satisfiability.
However, this is still restrictive compared to the more complex reasoning that KR formalisms provide.

Neuro-Symbolic Concept Learner \cite{mao19neuro} separates between visual perception and symbolic reasoning. It shows the data-efficiency by using only 10\% of the training data and achieving the state-of-the-art 98\% accuracy on CLEVR dataset.  
Our preliminary result $\nasp$ is similar in the sense that using symbolic reasoning, we could use fewer data to achieve a high accuracy. 

$\lpmln$ \cite{lee16weighted} serves as the foundation of our research. We started with extending $\lpmln$ with neural networks since existing $\lpmln$ parameter learning is already by the gradient descent method \cite{lee18weight} as used in the neural network training. However, there is a technical challenge: the previous parameter learning method does not scale up to be coupled with typical neural network training. This motivates us to consider a fragment of $\lpmln$ first, for which we design and implement an efficient parameter learning method. It turns out that this fragment is general enough to cover the ProbLog language as well as enjoys the expressiveness of full answer set programming language. 


\section{Goal of the Research}
The goal of our research is to develop some methods to integrate low-level perception with high-level reasoning so that, in inference tasks, reasoning can help identify perception mistakes that violate semantic constraints while, in learning tasks, a neural network not only learns from implicit correlations from the data but also from the explicit complex semantic constraints expressed by the rules. 

\begin{figure}[h!]
\vspace{-4mm}
\caption{Reasoning about relations among perceived objects}
\begin{center}
\includegraphics[width=10cm]{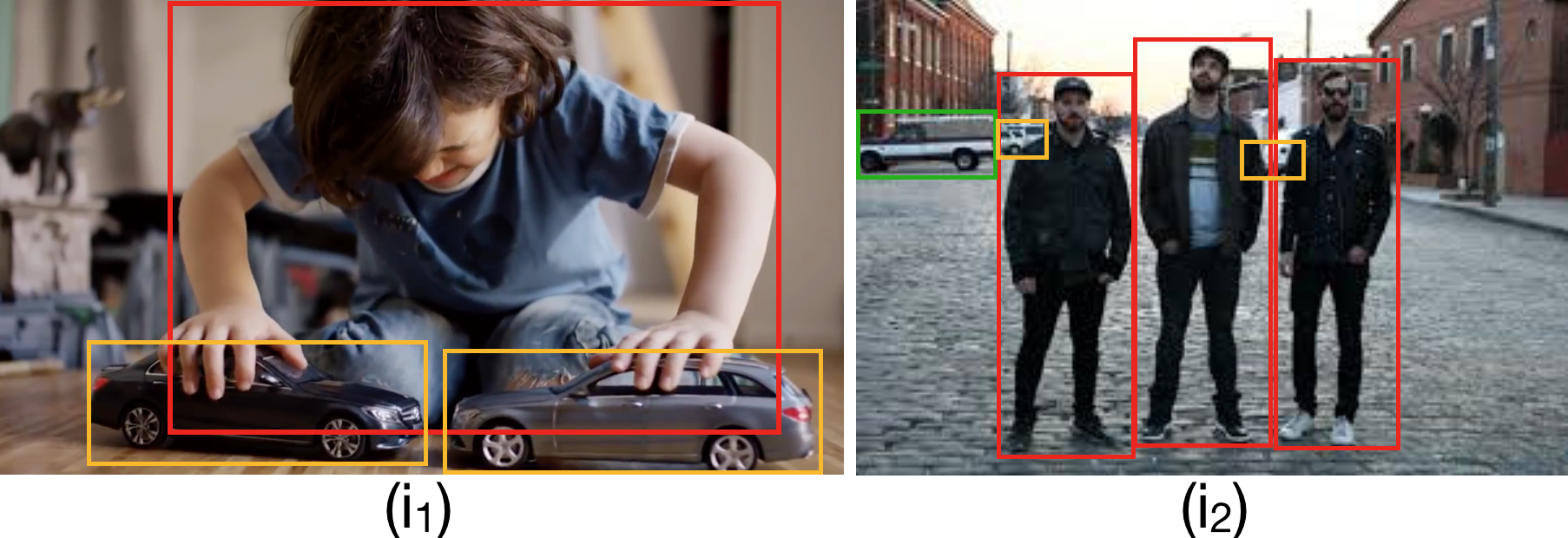}
\label{fig:yolo}
\end{center}
\vspace{-4mm}
\end{figure}

To achieve this goal, we investigate the integration of answer set programs with neural networks. We require that such integration should be not only expressive in representation but also efficient in computation. Besides, such integration should be able to apply reasoning in both inference and learning. For example, such integration should be able to reason about relations among perceived objects in Figure~\ref{fig:yolo} and tell that: the cars in ($i_1$) are toy cars since they are smaller than a person, while cars in ($i_2$) are real cars (by default) since there is no evidence to show they are smaller than those persons.

\section{Current Status of the Research}
To achieve our goal, we investigate and answer the following three research questions. This research is at a middle phase and the questions below are partially answered. 
\begin{enumerate}
\item How do we design a formalism that allows complex and high-level reasoning on information provided in vector space?
\newline
\textsl{We have one such design, $\nasp$ \cite{yang20neurasp}, that extends answer set programs with neural networks, using ideas from DeepProbLog \cite{manhaeve18deepproblog} to interface neural networks and ASP (i.e., treating the neural network output as the probability distribution over atomic facts in answer set programs), and using ideas from $\lpmln$ \cite{lee18weight} to design the probability and gradient computation under stable model semantics. 
\newline
We plan to design a new formalism that is less expressive but more scalable compared to $\nasp$ by doing all the reasoning within neural networks. 
To start with, we followed the ideas in \cite{xu18asemantic} to define a semantic loss using the logic rules. We represented implication rules by neural network regularizers for small domains including Nqueens problem, Sudoku, and Einstein's puzzle. We plan to design a general way to represent a CNF by a regularizer. We also plan to investigate into the rule forms that can be represented by a regularizer and then use these rule forms to design the new formalism.}

\item How do we implement such a formalism to make it as scalable as possible?
\newline
\textsl{The current implementation of $\nasp$ uses {\sc clingo} as its internal reasoning engine and uses {\sc PyTorch} to back-propagate the gradients from logic layer to neural networks, which yields the most scalable prototype among all of our trials so far and preserves the expressivity of ASP. 
Initially, we implemented the prototype of $\nasp$ where $\lpmln$ \cite{lee18weight} is used to compute the probability and gradient in $\nasp$ since $\lpmln$ is a probabilistic extension of ASP with well-defined probability and gradient computation under stable model semantics. However, the parameter learning method in $\lpmln$ does not scale up to be coupled with typical neural network training -- even for the MNIST addition example proposed in \cite{manhaeve18deepproblog}. To resolve this issue, we tried two approaches. First, we tested the idea of turning an answer set program into a Sentential Decision Diagram (SDD) but found that constructing such an SDD through the route ``ASP to CNF to SDD'' would take exponential time w.r.t. the number of atoms. There needs more research on SDD and possibly its variation that is more suitable for encoding answer set programs. Second, we tried to simplify $\lpmln$ to its fragment that is simple enough to have efficient computation and also expressive enough to capture all the ASP constructs. This leads to our last version of $\nasp$ to the date.
\newline
We are working on the prototype of the new formalism where logic rules are encoded in neural networks. We embedded implication rules in neural networks and successfully solved Nqueens problem, Sudoku, and Einstein's puzzle using neural networks only. 
We plan to embed CNFs in neural networks and implement a SAT solver using neural network only. Ultimately, we plan to embed answer set programs directly in neural networks. One challenge in this route is how to embed different constructs, such as negation, choice rules, and aggregation, in a neural network.
}
\item How do we evaluate and improve such a formalism?
\newline
\textsl{We evaluated $\nasp$ in \cite{yang20neurasp} w.r.t. the following domains: common-sense reasoning about image as in Figure~\ref{fig:yolo}, MNIST digit addition in \cite{manhaeve18deepproblog}, solving Sudoku in images in \cite{palm18recurrent}, and the shortest path problem in \cite{xu18asemantic}. We showed that $\nasp$ is very expressive and is able to help both neural network inference and training. We also showed that training with $\nasp$ still takes much more time than pure neural network training due to the internal use of a symbolic reasoning engine (i.e., {\sc clingo}). \newline
We plan to analyze the effect of $\nasp$ on other domains that require both perception and reasoning. The first domain in our agenda is visual question-answering while we limit our attention to those problems whose reasoning can be potentially represented in ASP. Some well-known datasets in this domain include NLVR2 \cite{suhr18corpus}, CLEVR \cite{johnson17clevr}, and CLEVRER \cite{yi19clevrer}. We plan to start with replacing the reasoning part in the existing work with $\nasp$ and analyze the pros and cons of applying $\nasp$ on those visual reasoning domains. 
We also plan to apply $\nasp$ to predict whether a vulnerability will be exploited so that the knowledge from domain experts can help neural network training.
What's more, since our preliminary results show that a neural network is not always trained better with more constraints, to know how to add constraints in real world problems, we also plan to analyze the effects of different constraints systematically.} 
\end{enumerate}

\section{Preliminary Results Accomplished}
We designed the syntax and the semantics of $\nasp$, and show how $\nasp$ can be useful for some tasks where both perception and high-level reasoning provided by answer set programs are required.

\subsection{NeurASP}
In \cite{yang20neurasp}, we present a simple extension of answer set programs by embracing neural networks.
Following the idea of DeepProbLog \cite{manhaeve18deepproblog}, by treating the neural network output as the probability distribution over atomic facts in answer set programs, the proposed $\nasp$ provides a simple and effective way to integrate  sub-symbolic and symbolic computation. 

In $\nasp$, a neural network $M$ is represented by a {\em neural atom} of the form 
\beq
   nn(m(e, t), \left[v_1, \dots, v_n \right]), 
\eeq{eq:nn-atom}
where
(i) $nn$ is a reserved keyword to denote a neural atom; 
(ii) $m$ is an identifier (symbolic name) of the neural network~$M$;
(iii) $t$ is a list of terms that serves as a ``pointer'' to an input tensor;  related to it, there is a mapping ${\bf D}$ (implemented by an external Python code) that turns $t$ into an input tensor; 
%
(iv) $v_1, \dots, v_n$ represent all $n$ possible outcomes of each of the $e$ random events.

Each neural atom \eqref{eq:nn-atom} introduces propositional atoms of the form $c\mvis v$, where $c\in\{m_1(t), \dots, m_e(t)\}$ and $v\in\{v_1,\dots,v_n\}$. The output of the neural network provides the probabilities of the introduced atoms.

\begin{example} \label{ex:digit}
Let $M_{digit}$ be a neural network that classifies an MNIST digit image. The input of $M_{digit}$ is (a tensor representation of) an image and the output is a matrix in $\mathbb{R}^{1\times 10}$. 
The neural network can be represented by the neural atom
\[
   nn( digit(1,d),\  [0,1,2,3,4,5,6,7,8,9]),
\]
which introduces propositional atoms $digit_1(d)\mvis 0$, $digit_1(d)\mvis 1$, $\dots$, $digit_1(d)\mvis 9$.
\end{example}

\BOCC
\begin{example}
Let $M_{sp}$ be another neural network for finding the shortest path in a graph with 24 edges. The input is a tensor that encodes the graph and the start/end nodes of the path, and the output is a matrix in $\mathbb{R}^{24\times 2}$.
This neural network can be represented by the neural atom
\[
    nn( sp(24,g),\  [\true, \false]) .
\]
\end{example}
\EOCC

A {\em $\nasp$ program} $\Pi$ is the union of $\Pi^{asp}$ and $\Pi^{nn}$, where $\Pi^{asp}$ is a set of propositional rules and $\Pi^{nn}$ is a set of neural atoms.  Let $\sigma^{nn}$ be the set of all atoms $m_i(t) \mvis v_j$ that is obtained from the neural atoms in $\Pi^{nn}$  as described above. We require that, in each rule $\i{Head}\ar\i{Body}$ in 
$\Pi^{asp}$, no atoms in $\sigma^{nn}$ appear in $\i{Head}$.

\BOCC
We could allow schematic variables into $\Pi$, which are understood in terms of grounding as in standard answer set programs.  We find it convenient to use rules of the form 
\beq
   nn(m(e, t), \left[v_1, \dots, v_n \right]) \ar \i{Body}
\eeq{eq:nn-rule}
where $\i{Body}$ is either identified by $\top$ or $\bot$ during grounding so that \eqref{eq:nn-rule} can be viewed as an abbreviation of multiple (variable-free) neural atoms~\eqref{eq:nn-atom}.

\begin{example}\label{ex:addition}
An example $\nasp$ program $\Pi_{digit}$ is as follows, where  $d_1$ and $d_2$ are terms representing two images. Each image is classified by neural network $M_{digit}$ as one of the values in $\{0,\dots,9 \}$. The addition of two digit-images is the sum of their values. 
\beq
{
\ba {l}
img(d_1). \hspace{4em}
img(d_2).\\[0.3em]
nn(digit(1, X), [0,1,2,3,4,5,6,7,8,9]) \leftarrow img(X).  \\[0.3em]
addition(A,B,N) \leftarrow
    digit_1(A) \mvis N_1, 
    digit_1(B) \mvis N_2, \\
\hspace{3.3cm} N=N_1+N_2.
\ea
}
\eeq{digit}
The neural network $M_{digit}$ outputs 10 probabilities for each image. The addition is applied once the digits are recognized and its probability is induced from the perception output as we explain in the next section. 
\end{example}
\EOCC


The semantics of $\nasp$ defines a {\em stable model} and its associated probability originating from the neural network output.
For any $\nasp$ program $\Pi$, we first obtain its ASP counterpart $\Pi'$ where each neural atom \eqref{eq:nn-atom} is replaced with the set of rules 
\[ 
\ba {l}
    \{m_i(t) \mvis v_1;~ \dots ~; m_i(t) \mvis v_n\} = 1   \ \ \ \ \text{for } i\in \{1,\dots e\}.
\ea
\]
%
We define the {\em stable models} of $\Pi$ as the stable models of $\Pi'$. %

\noindent{\bf Example~\ref{ex:digit} Continued~} The ASP counter-part of the neural atom in Example~\ref{ex:digit} is the following rule.
\[
\{digit_1(d)\mvis 0;~ \dots ~;  digit_1(d)\mvis 9\}=1.
\]

To define the probability of a stable model, we first define the probability of an atom $m_i(t)\mvis v_j$ in $\sigma^{nn}$. Recall that there is an external mapping ${\bf D}$ that turns $t$ into a specific input tensor ${\bf D}(t)$ of $M$. 
%
The probability of each atom $m_i(t) \mvis v_j$ is defined as $M({\bf D}(t))[i,j]$: 
\[
   P_{\Pi}(m_i(t) \mvis v_j) = M({\bf D}(t))[i,j] .\;
\]



The probability of a stable model $I$ of $\Pi$ is defined as the product of the probability of each atom $c=v$ in $I|_{\sigma^{nn}}$, divided by the number of stable models  of $\Pi$ that agree with $I|_{\sigma^{nn}}$ on $\sigma^{nn}$. 
That is, for any interpretation $I$, 
\[
P_{\Pi}(I) = \begin{cases}
\frac{\prod\limits_{c=v \in I|_{\sigma^{nn}}} P_{\Pi}(c=v)}{\i{Num}(I|_{\sigma^{nn}}, \Pi)}
 & \text{if $I$ is a stable model of $\Pi$;}\\
0 & \text{otherwise.}
\end{cases}
\]
where $I|_{\sigma^{nn}}$ denotes the projection of  $I$ onto $\sigma^{nn}$ and  $\i{Num}(I|_{\sigma^{nn}},\Pi)$ denotes the number of stable models of $\Pi$ that agree with $I|_{\sigma^{nn}}$ on $\sigma^{nn}$.

\subsection{Examples of NeurASP}
\begin{figure}[ht!]
\begin{center}
\caption{{\small $\nasp$ v.s. DeepProbLog v.s. Typical CNN}}
\includegraphics[scale=0.3]{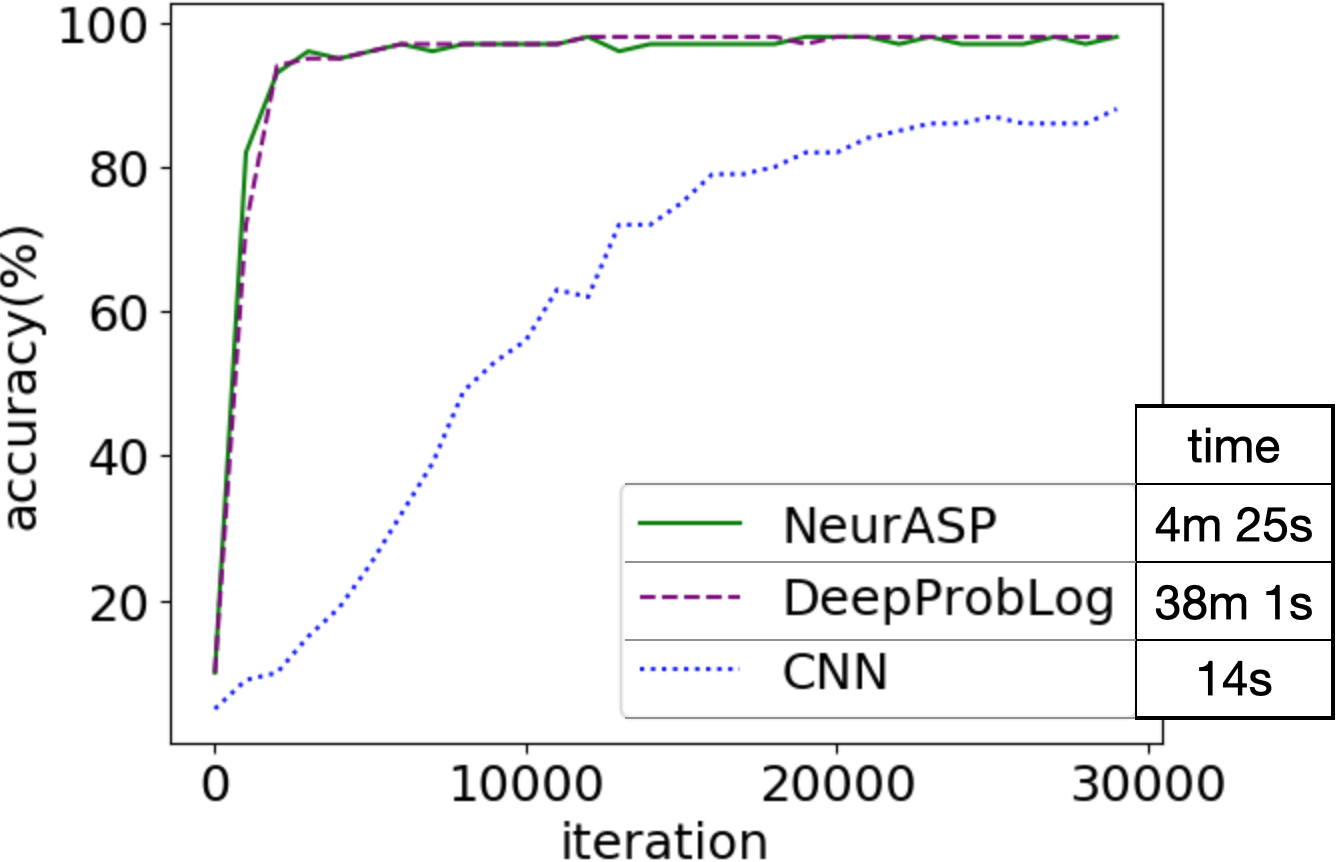}
\vspace{-0.4cm}
\label{fig:DeepLPMLNvsDeepProbLog}
\end{center}
\end{figure}

We show how $\nasp$ can be applied to the following 4 examples where both perception and high-level reasoning provided by answer set programs are required. 
\begin{itemize}
    \item {\bf MNIST Digit Addition~~} This is a simple example used in~\cite{manhaeve18deepproblog} to illustrate DeepProbLog's ability for both logical reasoning and deep learning. The task is, given a pair of digit images (MNIST) and their sum as the label, to let a neural network learn the digit classification of the input images. The $\nasp$ program is as follows.
\[
\ba {l}
img(d_1). ~~~img(d_2).\\[0.3em]
nn(digit(1, X), [0,1,2,3,4,5,6,7,8,9]) \leftarrow img(X).\\[0.3em]
addition(A,B,N) \leftarrow
    digit_1(A) \mvis N_1, 
    digit_1(B) \mvis N_2, 
    N=N_1+N_2.
\ea
\]
    Figure~\ref{fig:DeepLPMLNvsDeepProbLog} shows the accuracy on the test data after each training iteration. The method CNN denotes the baseline used in \cite{manhaeve18deepproblog} where a convolutional neural network (with more parameters) is trained to classify the concatenation of the two images into the 19 possible sums. 
    As we can see, the neural networks trained by $\nasp$ and DeepProbLog converge much faster than CNN and have almost the same accuracy at each iteration. 
    However, $\nasp$ spends much less time on training compared to DeepProbLog. The time reported is for one epoch (30,000 iterations in gradient descent).
    This is because DeepProbLog constructs an SDD at each iteration for each training instance (i.e., each pair of images). This example illustrates that generating many SDDs could be more time-consuming than enumerating stable models in $\nasp$ computation. In general, there is a trade-off between the two methods and other examples may show the opposite behavior.
\end{itemize}

\begin{itemize}
\item {\bf Commonsense Reasoning about Image~} 
We show how expressive reasoning originating from answer set programming, such as recursive definition and defaults can be used in $\nasp$ inference. We also show that reasoning in $\nasp$ can help identify perception mistakes that violate semantic constraints, which in turn can make perception more robust. 

Take the problem in Figure~\ref{fig:yolo} as an example. A neural network for object detection may return a bounding box and its classification ``car,'' but it may not be clear whether it is a real car or a toy car. The distinction can be made by applying reasoning about the relations with the surrounding objects and using commonsense knowledge. To reason about the size relationship among objects, we need to define a recursive definition of ``smaller than'' relationship.
\[
\ba {l}
smaller(cat, person). ~~smaller(person, car). ~~ smaller(person, truck).\\
smaller(X,Y) \leftarrow smaller(X,Z), smaller(Z,Y).
\ea
\]
We also need to use default reasoning to assert that, by default, we conclude the same size relationship as above between the objects in bounding boxes $B_1$ and $B_2$.
\[
\ba {l}
smaller(I, B_1, B_2) \leftarrow \i{not}\ \sneg smaller(I, B_1, B_2), 
label(I,B_1)\mvis L_1, label(I,B_2)\mvis L_2,
smaller(L_1,L_2).
\ea
\]
$\nasp$ allows the use of such expressive reasoning originating from answer set programming.

\item {\bf Solving Sudoku in Image~} We show that $\nasp$ alleviates the burden of neural networks when the constraints/knowledge are already given. Instead of building a large end-to-end neural network that learns to solve a Sudoku puzzle given as an image, we can let a neural network only do digit recognition and use ASP to find the solution of the recognized board. This makes the design of the neural network simpler and the
required training dataset much smaller. Also, the neural network may get confused if a digit next to $1$ in the same row is $1$ or $2$, but the reasoner can conclude that it cannot be $1$ by applying the constraints for Sudoku. 
What's more, when we need to solve a variant of Sudoku, such as Anti-knight Sudoku, the modification is much simpler than training another large neural network from scratch to solve the new puzzle. Indeed, one can use the same pre-trained neural network and only need to add a rule in the $\nasp$ program saying that ``no number repeats at a knight move''.

\item {\bf Learning Shortest Path~} We show how expressive reasoning originating from ASP, such as recursive definition, aggregates, and weak constraints, can be used in $\nasp$ learning. We aim at training a neural network to find a shortest path in a $4\times 4$ grid with missing edges. Such a neural network can be used to simulate a plain ASP program to solve the same problem using less inference time.
The representation of the shortest path problem in $\nasp$ would be almost the same as a {\sc clingo} program and as simple as follows.
\begin{lstlisting}
nn(sp(24, g), [true, false]).
% if edge 1 in graph g is selected in the shortest path, 
% then there is an edge between node 0 and node 1
sp(0,1) :- sp(1,g,true).    
...      
sp(X,Y) :- sp(Y,X).

% [nr] No removed edges should be predicted
:- sp(X,g,true), removed(X).

% [p] (aggregates) Prediction must form a simple path, i.e.,
%                  the degree of each node must be either 0 or 2
:- X=0..15, #count{Y: sp(X,Y)} = 1.
:- X=0..15, #count{Y: sp(X,Y)} >= 3.

% [r] (recursive definition) Every 2 nodes in the prediction must be reachable
reachable(X,Y) :- sp(X,Y).
reachable(X,Y) :- reachable(X,Z), sp(Z,Y).
:- sp(X,A), sp(Y,B), not reachable(X,Y).

% [o] (weak constraint) Predicted path should contain least edges 
:~ sp(X,g,true). [1, X]
\end{lstlisting}
In this experiment, we trained the same neural network model $M_{sp}$ as in \cite{xu18asemantic}, a 5-layer Multi-Layer Perceptron (MLP), but with 4 different settings: (i) MLP only; (ii) together with $\nasp$ with the simple-path constraint {\bf (p)} (which is the only constraint used in \cite{xu18asemantic}); 
\footnote{A path is {\em simple} if every node in it other than the source and the destination has only 1 incoming and only 1 outgoing edge.}
(iii) together with $\nasp$ with simple-path, reachability, and optimization constraints {\bf (p-r-o)}; and (iv) together with $\nasp$ with all 4 constraints {\bf (p-r-o-nr)}. 
\footnote{Other combinations are either meaningless (e.g., {\bf o}) or having similar results (e.g. {\bf p-r} is similar to {\bf p}).}

Table~\ref{tb:sp:Accuracy on Test Data} shows, after 500 epochs of training, the percentage of the predictions on the test data that satisfy each of the constraints {\bf p}, {\bf r}, and {\bf nr},  the path constraint (i.e., {\bf p-r}), the shortest path constraint (i.e., {\bf p-r-o-nr}), and the accuracy w.r.t. the ground truth.
As we can see, $\nasp$ helps to train the same neural network such that it's more likely to satisfy the constraints. Besides, the last column shows that a neural network is not always trained better with more constraints

\vspace{-0.2cm}
\begin{table}[ht]
\caption{Shortest Path: Accuracy on Test Data: columns denote MLPs trained with different rules; each row represents the percentage of predictions that satisfy the constraints}
\centering
{\small 
\begin{tabular} {c| c| c| c| c}
\hline
\hline
Predictions & {MLP Only} & MLP & MLP & MLP\\
satisfying &  & (p) & (p-r-o) & (p-r-o-nr)\\
\hline
p & 28.3\% & 96.6\% & {\bf 100}\% & 30.1\%\\
r & 88.5\% & {\bf 100}\% &  {\bf 100}\% & 87.3\% \\
nr & 32.9\% & 36.3\% &  45.7\% & {\bf 70.5}\%\\
p-r &28.3\% & 96.6\% & {\bf 100}\% & 30.1\%\\
p-r-o-nr & 23.0\% & 33.2\% & {\bf 45.7}\% & 24.2\% \\
{\sl label (ground truth)} & 22.4\% & 28.9\% & {\bf 40.1}\% & 22.7\%\\
\hline
\hline
\end{tabular}
\label{tb:sp:Accuracy on Test Data}
}
\end{table}

\end{itemize}

\section{Open Issues and Expected Achievements}
The biggest issue we are encountering now is that training with $\nasp$ implementation still takes much more time than pure neural network training due to the internal use of a symbolic reasoning engine (i.e., {\sc clingo} in our case). With our recent experience on encoding logic directly in neural networks, we expect that the scalability issue will be resolved by embedding ASP (or possibly a fragment of ASP) directly in neural networks.
We expect to design a new formalism whose rules can be turned into neural network regularizers. We also expect to implement a prototype for the new formalism and apply it to the domains that $\nasp$ was applied to so that we can compare and analyze the pros and cons of both approach.

\nocite{*}
\bibliographystyle{eptcs}
\bibliography{zhun}

\begin{thebibliography}{10}
\providecommand{\bibitemdeclare}[2]{}
\providecommand{\surnamestart}{}
\providecommand{\surnameend}{}
\providecommand{\urlprefix}{Available at }
\providecommand{\url}[1]{\texttt{#1}}
\providecommand{\href}[2]{\texttt{#2}}
\providecommand{\urlalt}[2]{\href{#1}{#2}}
\providecommand{\doi}[1]{doi:\urlalt{http://dx.doi.org/#1}{#1}}
\providecommand{\bibinfo}[2]{#2}

\bibitemdeclare{inproceedings}{andreas16learning}
\bibitem{andreas16learning}
\bibinfo{author}{Jacob \surnamestart Andreas\surnameend},
  \bibinfo{author}{Marcus \surnamestart Rohrbach\surnameend},
  \bibinfo{author}{Trevor \surnamestart Darrell\surnameend} \&
  \bibinfo{author}{Dan \surnamestart Klein\surnameend} (\bibinfo{year}{2016}):
  \emph{\bibinfo{title}{Learning to Compose Neural Networks for Question
  Answering}}.
\newblock In: {\sl \bibinfo{booktitle}{Proceedings of the 2016 Annual
  Conference of the North American Chapter of the Association for Computational
  Linguistics: Human Language Technologies}}, pp. \bibinfo{pages}{1545--1554},
  \doi{10.18653/v1/n16-1181}.

\bibitemdeclare{proceedings}{baader03handbook}
\bibitem{baader03handbook}
\bibinfo{editor}{Franz \surnamestart Baader\surnameend}, \bibinfo{editor}{Diego
  \surnamestart Calvanese\surnameend}, \bibinfo{editor}{Deborah~L.
  \surnamestart McGuinness\surnameend}, \bibinfo{editor}{Daniele \surnamestart
  Nardi\surnameend} \& \bibinfo{editor}{Peter~F. \surnamestart
  Patel-Schneider\surnameend}, editors (\bibinfo{year}{2003}):
  \emph{\bibinfo{title}{The Description Logic Handbook: Theory, Implementation,
  and Applications}}. \bibinfo{publisher}{Cambridge University Press}.

\bibitemdeclare{article}{bre11}
\bibitem{bre11}
\bibinfo{author}{Gerhard \surnamestart Brewka\surnameend},
  \bibinfo{author}{Ilkka \surnamestart Niemel\"{a}\surnameend} \&
  \bibinfo{author}{Miroslaw \surnamestart Truszczynski\surnameend}
  (\bibinfo{year}{2011}): \emph{\bibinfo{title}{Answer Set Programming at a
  Glance}}.
\newblock {\sl \bibinfo{journal}{Communications of the ACM}}
  \bibinfo{volume}{54(12)}, pp. \bibinfo{pages}{92--103},
  \doi{10.1145/2043174.2043195}.

\bibitemdeclare{article}{calimeri20asp}
\bibitem{calimeri20asp}
\bibinfo{author}{Francesco \surnamestart Calimeri\surnameend},
  \bibinfo{author}{Wolfgang \surnamestart Faber\surnameend},
  \bibinfo{author}{Martin \surnamestart Gebser\surnameend},
  \bibinfo{author}{Giovambattista \surnamestart Ianni\surnameend},
  \bibinfo{author}{Roland \surnamestart Kaminski\surnameend},
  \bibinfo{author}{Thomas \surnamestart Krennwallner\surnameend},
  \bibinfo{author}{Nicola \surnamestart Leone\surnameend},
  \bibinfo{author}{Marco \surnamestart Maratea\surnameend},
  \bibinfo{author}{Francesco \surnamestart Ricca\surnameend} \&
  \bibinfo{author}{Torsten \surnamestart Schaub\surnameend}
  (\bibinfo{year}{2020}): \emph{\bibinfo{title}{{ASP-C}ore-2 input language
  format}}.
\newblock {\sl \bibinfo{journal}{Theory and Practice of Logic Programming}}
  \bibinfo{volume}{20}(\bibinfo{number}{2}), pp. \bibinfo{pages}{294--309},
  \doi{10.1017/S1471068419000450}.

\bibitemdeclare{article}{kathryn18tensorlog}
\bibitem{kathryn18tensorlog}
\bibinfo{author}{William~W \surnamestart Cohen\surnameend},
  \bibinfo{author}{Fan \surnamestart Yang\surnameend} \&
  \bibinfo{author}{Kathryn~Rivard \surnamestart Mazaitis\surnameend}
  (\bibinfo{year}{2018}): \emph{\bibinfo{title}{TensorLog: Deep Learning Meets
  Probabilistic Databases}}.
\newblock {\sl \bibinfo{journal}{Journal of Artificial Intelligence Research}}
  \bibinfo{volume}{1}, pp. \bibinfo{pages}{1--15}.

\bibitemdeclare{inproceedings}{donadello17logic}
\bibitem{donadello17logic}
\bibinfo{author}{Ivan \surnamestart Donadello\surnameend},
  \bibinfo{author}{Luciano \surnamestart Serafini\surnameend} \&
  \bibinfo{author}{Artur~D'Avila \surnamestart Garcez\surnameend}
  (\bibinfo{year}{2017}): \emph{\bibinfo{title}{Logic tensor networks for
  semantic image interpretation}}.
\newblock In: {\sl \bibinfo{booktitle}{Proceedings of the 26th International
  Joint Conference on Artificial Intelligence}}, \bibinfo{organization}{AAAI
  Press}, pp. \bibinfo{pages}{1596--1602}, \doi{10.24963/ijcai.2017/221}.

\bibitemdeclare{inproceedings}{johnson17clevr}
\bibitem{johnson17clevr}
\bibinfo{author}{Justin \surnamestart Johnson\surnameend},
  \bibinfo{author}{Bharath \surnamestart Hariharan\surnameend},
  \bibinfo{author}{Laurens \surnamestart van~der Maaten\surnameend},
  \bibinfo{author}{Li~\surnamestart Fei-Fei\surnameend},
  \bibinfo{author}{C~\surnamestart Lawrence~Zitnick\surnameend} \&
  \bibinfo{author}{Ross \surnamestart Girshick\surnameend}
  (\bibinfo{year}{2017}): \emph{\bibinfo{title}{Clevr: A diagnostic dataset for
  compositional language and elementary visual reasoning}}.
\newblock In: {\sl \bibinfo{booktitle}{Proceedings of the IEEE Conference on
  Computer Vision and Pattern Recognition}}, pp. \bibinfo{pages}{2901--2910},
  \doi{10.1109/CVPR.2017.215}.

\bibitemdeclare{inproceedings}{kazemi18relnn}
\bibitem{kazemi18relnn}
\bibinfo{author}{Seyed~Mehran \surnamestart Kazemi\surnameend} \&
  \bibinfo{author}{David \surnamestart Poole\surnameend}
  (\bibinfo{year}{2018}): \emph{\bibinfo{title}{RelNN: A deep neural model for
  relational learning}}.
\newblock In: {\sl \bibinfo{booktitle}{Proceedings of the 32nd AAAI Conference
  on Artificial Intelligence}}.

\bibitemdeclare{inproceedings}{kipf17semi}
\bibitem{kipf17semi}
\bibinfo{author}{Thomas~N. \surnamestart Kipf\surnameend} \&
  \bibinfo{author}{Max \surnamestart Welling\surnameend}
  (\bibinfo{year}{2017}): \emph{\bibinfo{title}{Semi-Supervised Classification
  with Graph Convolutional Networks}}.
\newblock In: {\sl \bibinfo{booktitle}{Proceedings of the 5th International
  Conference on Learning Representations, {ICLR} 2017}}.

\bibitemdeclare{inproceedings}{kisa14probabilistic}
\bibitem{kisa14probabilistic}
\bibinfo{author}{Doga \surnamestart Kisa\surnameend}, \bibinfo{author}{Guy
  \surnamestart Van~den Broeck\surnameend}, \bibinfo{author}{Arthur
  \surnamestart Choi\surnameend} \& \bibinfo{author}{Adnan \surnamestart
  Darwiche\surnameend} (\bibinfo{year}{2014}):
  \emph{\bibinfo{title}{Probabilistic sentential decision diagrams}}.
\newblock In: {\sl \bibinfo{booktitle}{Fourteenth International Conference on
  the Principles of Knowledge Representation and Reasoning}}.

\bibitemdeclare{inproceedings}{lee16weighted}
\bibitem{lee16weighted}
\bibinfo{author}{Joohyung \surnamestart Lee\surnameend} \&
  \bibinfo{author}{Yi~\surnamestart Wang\surnameend} (\bibinfo{year}{2016}):
  \emph{\bibinfo{title}{Weighted Rules under the Stable Model Semantics}}.
\newblock In: {\sl \bibinfo{booktitle}{Proceedings of International Conference
  on Principles of Knowledge Representation and Reasoning (KR)}}, pp.
  \bibinfo{pages}{145--154}.

\bibitemdeclare{inproceedings}{lee18weight}
\bibitem{lee18weight}
\bibinfo{author}{Joohyung \surnamestart Lee\surnameend} \&
  \bibinfo{author}{Yi~\surnamestart Wang\surnameend} (\bibinfo{year}{2018}):
  \emph{\bibinfo{title}{Weight Learning in a Probabilistic Extension of Answer
  Set Programs}}.
\newblock In: {\sl \bibinfo{booktitle}{Proceedings of International Conference
  on Principles of Knowledge Representation and Reasoning (KR)}}, pp.
  \bibinfo{pages}{22--31}.

\bibitemdeclare{inproceedings}{lee17lpmln}
\bibitem{lee17lpmln}
\bibinfo{author}{Joohyung \surnamestart Lee\surnameend} \&
  \bibinfo{author}{Zhun \surnamestart Yang\surnameend} (\bibinfo{year}{2017}):
  \emph{\bibinfo{title}{{L}{P}{M}{L}{N}, Weak Constraints, and {P}-log}}.
\newblock In: {\sl \bibinfo{booktitle}{Proceedings of the AAAI Conference on
  Artificial Intelligence (AAAI)}}, pp. \bibinfo{pages}{1170--1177}.

\bibitemdeclare{inproceedings}{lierler04cmodels}
\bibitem{lierler04cmodels}
\bibinfo{author}{Yuliya \surnamestart Lierler\surnameend} \&
  \bibinfo{author}{Marco \surnamestart Maratea\surnameend}
  (\bibinfo{year}{2004}): \emph{\bibinfo{title}{Cmodels-2: SAT-based answer set
  solver enhanced to non-tight programs}}.
\newblock In: {\sl \bibinfo{booktitle}{Proceedings of International Conference
  on Logic Programming and NonMonotonic Reasoning}},
  \bibinfo{organization}{Springer}, pp. \bibinfo{pages}{346--350},
  \doi{10.1007/978-3-540-24609-1\_32}.

\bibitemdeclare{inproceedings}{lif08}
\bibitem{lif08}
\bibinfo{author}{Vladimir \surnamestart Lifschitz\surnameend}
  (\bibinfo{year}{2008}): \emph{\bibinfo{title}{What Is Answer Set
  Programming?}}
\newblock In: {\sl \bibinfo{booktitle}{Proceedings of the AAAI Conference on
  Artificial Intelligence}}, \bibinfo{publisher}{MIT Press}, pp.
  \bibinfo{pages}{1594--1597}.

\bibitemdeclare{inproceedings}{lin19kagnet}
\bibitem{lin19kagnet}
\bibinfo{author}{Bill~Yuchen \surnamestart Lin\surnameend},
  \bibinfo{author}{Xinyue \surnamestart Chen\surnameend},
  \bibinfo{author}{Jamin \surnamestart Chen\surnameend} \&
  \bibinfo{author}{Xiang \surnamestart Ren\surnameend} (\bibinfo{year}{2019}):
  \emph{\bibinfo{title}{KagNet: Knowledge-Aware Graph Networks for Commonsense
  Reasoning}}.
\newblock In: {\sl \bibinfo{booktitle}{Proceedings of the 2019 Conference on
  Empirical Methods in Natural Language Processing and the 9th International
  Joint Conference on Natural Language Processing (EMNLP-IJCNLP)}}, pp.
  \bibinfo{pages}{2822--2832}, \doi{10.18653/v1/D19-1282}.

\bibitemdeclare{inproceedings}{manhaeve18deepproblog}
\bibitem{manhaeve18deepproblog}
\bibinfo{author}{Robin \surnamestart Manhaeve\surnameend},
  \bibinfo{author}{Sebastijan \surnamestart Dumancic\surnameend},
  \bibinfo{author}{Angelika \surnamestart Kimmig\surnameend},
  \bibinfo{author}{Thomas \surnamestart Demeester\surnameend} \&
  \bibinfo{author}{Luc \surnamestart De~Raedt\surnameend}
  (\bibinfo{year}{2018}): \emph{\bibinfo{title}{Deepproblog: Neural
  probabilistic logic programming}}.
\newblock In: {\sl \bibinfo{booktitle}{Proceedings of Advances in Neural
  Information Processing Systems}}, pp. \bibinfo{pages}{3749--3759}.

\bibitemdeclare{inproceedings}{mao19neuro}
\bibitem{mao19neuro}
\bibinfo{author}{Jiayuan \surnamestart Mao\surnameend}, \bibinfo{author}{Chuang
  \surnamestart Gan\surnameend}, \bibinfo{author}{Pushmeet \surnamestart
  Kohli\surnameend}, \bibinfo{author}{Joshua~B. \surnamestart
  Tenenbaum\surnameend} \& \bibinfo{author}{Jiajun \surnamestart Wu\surnameend}
  (\bibinfo{year}{2019}): \emph{\bibinfo{title}{The neuro-symbolic concept
  learner: interpreting scenes, words, and sentences from natural
  supervision}}.
\newblock In: {\sl \bibinfo{booktitle}{Proceedings of International Conference
  on Learning Representations}}.

\bibitemdeclare{inproceedings}{palm18recurrent}
\bibitem{palm18recurrent}
\bibinfo{author}{Rasmus \surnamestart Palm\surnameend}, \bibinfo{author}{Ulrich
  \surnamestart Paquet\surnameend} \& \bibinfo{author}{Ole \surnamestart
  Winther\surnameend} (\bibinfo{year}{2018}): \emph{\bibinfo{title}{Recurrent
  relational networks}}.
\newblock In: {\sl \bibinfo{booktitle}{Proceedings of Advances in Neural
  Information Processing Systems}}, pp. \bibinfo{pages}{3368--3378}.

\bibitemdeclare{book}{pearl00causality}
\bibitem{pearl00causality}
\bibinfo{author}{Judea \surnamestart Pearl\surnameend} (\bibinfo{year}{2000}):
  \emph{\bibinfo{title}{Causality: models, reasoning and inference}}.
\newblock \bibinfo{volume}{29}, \bibinfo{publisher}{Cambridge Univ Press}.

\bibitemdeclare{article}{rei80}
\bibitem{rei80}
\bibinfo{author}{Raymond \surnamestart Reiter\surnameend}
  (\bibinfo{year}{1980}): \emph{\bibinfo{title}{A logic for default
  reasoning}}.
\newblock {\sl \bibinfo{journal}{Artificial Intelligence}}
  \bibinfo{volume}{13}, pp. \bibinfo{pages}{81--132},
  \doi{10.1016/0004-3702(80)90014-4}.

\bibitemdeclare{inproceedings}{rocktaschel17end}
\bibitem{rocktaschel17end}
\bibinfo{author}{Tim \surnamestart Rockt{\"a}schel\surnameend} \&
  \bibinfo{author}{Sebastian \surnamestart Riedel\surnameend}
  (\bibinfo{year}{2017}): \emph{\bibinfo{title}{End-to-end differentiable
  proving}}.
\newblock In: {\sl \bibinfo{booktitle}{Proceedings of Advances in Neural
  Information Processing Systems}}, pp. \bibinfo{pages}{3788--3800}.

\bibitemdeclare{inproceedings}{selsam18learning}
\bibitem{selsam18learning}
\bibinfo{author}{Daniel \surnamestart Selsam\surnameend},
  \bibinfo{author}{Matthew \surnamestart Lamm\surnameend},
  \bibinfo{author}{Benedikt \surnamestart B{\"{u}}nz\surnameend},
  \bibinfo{author}{Percy \surnamestart Liang\surnameend},
  \bibinfo{author}{Leonardo \surnamestart de~Moura\surnameend} \&
  \bibinfo{author}{David~L. \surnamestart Dill\surnameend}
  (\bibinfo{year}{2019}): \emph{\bibinfo{title}{Learning a {SAT} Solver from
  Single-Bit Supervision}}.
\newblock In: {\sl \bibinfo{booktitle}{Proceedings of the 7th International
  Conference on Learning Representations (ICLR)}}.

\bibitemdeclare{inproceedings}{sourek15lifted}
\bibitem{sourek15lifted}
\bibinfo{author}{Gustav \surnamestart {\v{S}}ourek\surnameend},
  \bibinfo{author}{Vojtech \surnamestart Aschenbrenner\surnameend},
  \bibinfo{author}{Filip \surnamestart {\v{Z}}elezny\surnameend} \&
  \bibinfo{author}{Ond{\v{r}}ej \surnamestart Ku{\v{z}}elka\surnameend}
  (\bibinfo{year}{2015}): \emph{\bibinfo{title}{Lifted relational neural
  networks}}.
\newblock In: {\sl \bibinfo{booktitle}{Proceedings of the 2015th International
  Conference on Cognitive Computation: Integrating Neural and Symbolic
  Approaches-Volume 1583}}, \bibinfo{organization}{CEUR-WS. org}, pp.
  \bibinfo{pages}{52--60}.

\bibitemdeclare{inproceedings}{suhr18corpus}
\bibitem{suhr18corpus}
\bibinfo{author}{Alane \surnamestart Suhr\surnameend},
  \bibinfo{author}{Stephanie \surnamestart Zhou\surnameend},
  \bibinfo{author}{Ally \surnamestart Zhang\surnameend}, \bibinfo{author}{Iris
  \surnamestart Zhang\surnameend}, \bibinfo{author}{Huajun \surnamestart
  Bai\surnameend} \& \bibinfo{author}{Yoav \surnamestart Artzi\surnameend}
  (\bibinfo{year}{2019}): \emph{\bibinfo{title}{A Corpus for Reasoning about
  Natural Language Grounded in Photographs}}.
\newblock In: {\sl \bibinfo{booktitle}{Proceedings of the 57th Conference of
  the Association for Computational Linguistics (ACL)}}, pp.
  \bibinfo{pages}{6418--6428}, \doi{10.18653/v1/p19-1644}.

\bibitemdeclare{inproceedings}{xu18asemantic}
\bibitem{xu18asemantic}
\bibinfo{author}{Jingyi \surnamestart Xu\surnameend}, \bibinfo{author}{Zilu
  \surnamestart Zhang\surnameend}, \bibinfo{author}{Tal \surnamestart
  Friedman\surnameend}, \bibinfo{author}{Yitao \surnamestart Liang\surnameend}
  \& \bibinfo{author}{Guy \surnamestart Van~den Broeck\surnameend}
  (\bibinfo{year}{2018}): \emph{\bibinfo{title}{A Semantic Loss Function for
  Deep Learning with Symbolic Knowledge}}.
\newblock In: {\sl \bibinfo{booktitle}{Proceedings of the 35th International
  Conference on Machine Learning (ICML)}}.
\newblock \urlprefix\url{http://starai.cs.ucla.edu/papers/XuICML18.pdf}.

\bibitemdeclare{inproceedings}{yang20neurasp}
\bibitem{yang20neurasp}
\bibinfo{author}{Zhun \surnamestart Yang\surnameend}, \bibinfo{author}{Adam
  \surnamestart Ishay\surnameend} \& \bibinfo{author}{Joohyung \surnamestart
  Lee\surnameend} (\bibinfo{year}{2020}): \emph{\bibinfo{title}{NeurASP:
  Embracing Neural Networks into Answer Set Programming}}.
\newblock In: {\sl \bibinfo{booktitle}{Proceedings of International Joint
  Conference on Artificial Intelligence (IJCAI)}}, pp.
  \bibinfo{pages}{1755--1762}, \doi{10.1017/S1471068419000450}.

\bibitemdeclare{inproceedings}{yi19clevrer}
\bibitem{yi19clevrer}
\bibinfo{author}{Kexin \surnamestart Yi\surnameend}, \bibinfo{author}{Chuang
  \surnamestart Gan\surnameend}, \bibinfo{author}{Yunzhu \surnamestart
  Li\surnameend}, \bibinfo{author}{Pushmeet \surnamestart Kohli\surnameend},
  \bibinfo{author}{Jiajun \surnamestart Wu\surnameend},
  \bibinfo{author}{Antonio \surnamestart Torralba\surnameend} \&
  \bibinfo{author}{Joshua~B. \surnamestart Tenenbaum\surnameend}
  (\bibinfo{year}{2020}): \emph{\bibinfo{title}{{CLEVRER:} Collision Events for
  Video Representation and Reasoning}}.
\newblock In: {\sl \bibinfo{booktitle}{Proceedings of the 8th International
  Conference on Learning Representations (ICLR)}}.

\end{thebibliography}
\end{document}